%% file: main.tex
\documentclass{article}
\usepackage[english]{babel}
\usepackage[numbers]{natbib}
\usepackage{float}
\usepackage{gensymb}
\usepackage[preprint]{neurips_2025}
\usepackage{setspace}
\usepackage{booktabs}
\usepackage{placeins}
\usepackage{tabularx}
\usepackage{makecell}
\usepackage{multirow}
\usepackage{amsmath}
\usepackage{graphicx}
\usepackage[skip=10pt, belowskip=10pt]{caption}
\usepackage[utf8]{inputenc} 
\usepackage[T1]{fontenc}    
\usepackage{xcolor}         
\usepackage{hyperref}       
\usepackage{url}            
\usepackage{amsfonts}       
\usepackage{nicefrac}       
\usepackage{microtype}      
\usepackage{csquotes}
\usepackage{algorithm}
\usepackage{algpseudocode}
\usepackage{subcaption}
\usepackage{mwe}
\usepackage{cleveref}
\newsavebox{\subfigTall}
\newsavebox{\subfigShort}
\title{Self-Supervised Animal Identification for Long Videos}
\author{Xuyang Fang, Sion Hannuna, Edwin Simpson, Neill Campbell\\
University of Bristol\\
\{xf16910, sh1670, Neill.Campbell, edwin.simpson\}@bristol.ac.uk
}

\begin{document}
\maketitle



\begin{abstract}
Identifying individual animals in long-duration videos is essential for 
behavioral ecology, wildlife monitoring, and livestock management. Traditional 
methods require extensive manual annotation, while existing self-supervised 
approaches are computationally demanding and ill-suited for long sequences due 
to memory constraints and temporal error propagation. We introduce a highly 
efficient, self-supervised method that reframes animal identification as a 
global clustering task rather than a sequential tracking problem. Our approach 
assumes a known, fixed number of individuals within a single video---a common 
scenario in practice---and requires only bounding box detections and the total 
count. By sampling pairs of frames, using a frozen pre-trained backbone, and 
employing a self-bootstrapping mechanism with the Hungarian algorithm for 
in-batch pseudo-label assignment, our method learns discriminative features 
without identity labels. We adapt a Binary Cross Entropy loss from vision-language
models, enabling state-of-the-art accuracy ($>$97\%) while consuming less than 
1 GB of GPU memory per batch--an order of magnitude less than standard contrastive 
methods. Evaluated on challenging real-world datasets (3D-POP pigeons and 
8-calves feeding videos), our framework matches or surpasses supervised 
baselines trained on over 1,000 labeled frames, effectively removing the 
manual annotation bottleneck. This work enables practical, high-accuracy 
animal identification on consumer-grade hardware, with broad applicability 
in resource-constrained research settings. All code written for this paper
are \href{https://huggingface.co/datasets/tonyFang04/8-calves}{here}.
\end{abstract}

\section{Introduction}

\input{introduction.tex}

\section{Method}

\input{method.tex}

\FloatBarrier
\section{Experimental Details}

\input{experimental_details.tex}

\FloatBarrier
\section{Results}

\input{results.tex}

\FloatBarrier
\section{Related Work}

\input{related_work.tex}

\section{Conclusion}

\input{conclusion.tex}

\FloatBarrier
\bibliographystyle{IEEEtran}
\bibliography{reference}


\end{document}

%% file: introduction.tex
Identifying individual animals across long-duration video sequences 
is a fundamental challenge in behavioral ecology, wildlife monitoring, 
and livestock management. The ability to consistently recognize individuals 
without manual intervention enables researchers to track behavior, social 
interactions, and health over time, providing critical insights into animal 
welfare and population dynamics. Traditional approaches often rely on 
supervised learning, which requires extensive manual annotation of identity 
labels—a tedious, costly, and often impractical process for large video datasets. 
While recent advances in self-supervised learning offer promising alternatives 
by learning representations without explicit labels, these methods are typically 
designed for general-purpose visual representation learning and are ill-suited 
to the specific constraints of animal identification in long videos.

Two primary challenges hinder the application of existing self-supervised 
frameworks to this domain. First, computational and memory constraints limit 
scalability: methods such as SimCLR and MoCo require large batch sizes or 
memory banks to learn effective representations, often consuming over 10GB 
of GPU memory per batch—a prohibitive requirement for many research settings. 
Second, temporal error propagation plagues sequential tracking paradigms: 
even state-of-the-art tracking methods (e.g., 3D-MuPPET) rely on frame-by-frame 
association, where a single identity switch can propagate indefinitely in long 
sequences, degrading performance over time. These limitations are particularly 
acute in videos lasting hours or days, where manual correction is infeasible 
and computational resources are often constrained.

Our work addresses a specific but common and valuable scenario: the 
identification of a known, fixed number of individual animals within a single 
video. This assumption holds true in many practical settings, such as monitoring 
a defined group of livestock in a pen, a specific colony of laboratory animals, 
or a set number of birds at a feeder. Under this assumption, we reframe the 
animal identification problem as a global clustering task rather than a 
sequential tracking problem. We introduce a resource-efficient, self-supervised 
method that learns to distinguish individuals without any identity labels, relying 
only on pre-existing bounding box detections and the prior knowledge of the 
total count of distinct individuals.

Our approach samples pairs of frames from a video, extracts animal crops, 
and applies augmentations to create multiple views. Using a pre-trained 
backbone network, we compute feature vectors and construct a similarity 
matrix across all augmented instances. We used the Hungarian algorithm within 
each training batch to dynamically assign positive pairs based on feature 
similarity, effectively generating pseudo-labels that guide representation 
learning. This self-bootstrapping mechanism allows the network to learn 
discriminative features specific to the fixed set of animal identities, even 
when initialized with generic pre-trained weights. After training, a single 
pass over all video frames produces feature embeddings that are clustered 
using K-Means, with the number of clusters set to the known population size.

Designed with minimal memory consumption as a core objective, our method 
samples only two frames per batch, allows for a frozen backbone, and employs 
efficient loss functions--including a Binary Cross Entropy (BCE) loss adapted 
from vision-language models like SigLIP\cite{siglip}. This reduces memory 
usage to below 1GB per batch while achieving state-of-the-art accuracy. 
We evaluate our approach on challenging real-world datasets, including 
the 3D-POP pigeon dataset and an 8-calves feeding video, demonstrating 
robust performance ($>$97\% accuracy) that matches or exceeds supervised 
baselines trained on over 1,000 labeled frames.

While our approach shares the use of random frame sampling and the Hungarian 
algorithm for in-batch pseudo-label assignment with recent multi-object 
tracking work\cite{Plaen_2024}, our key contributions lie in the specific 
design choices that enable high efficiency and applicability to
single-video animal identification:

\begin{enumerate}
    \item \textbf{Single-video, single-species optimization:} Unlike methods 
    that sample from multiple videos to maximize negative diversity, we sample 
    frames solely from the target video. This reduces computational overhead and 
    memory footprint while being sufficient for learning discriminative features 
    within a closed population.

    \item \textbf{Frozen backbone compatibility:} Our method functions effectively 
    even when the feature extraction backbone is frozen, requiring only a lightweight 
    projection head to be trained. This reduces memory usage by approximately 60\% 
    while maintaining high accuracy, leveraging the transfer learning capabilities 
    of pre-trained models.

    \item \textbf{Minimalist loss formulation:} We demonstrate that a single loss 
    function--either a Supervised Contrastive Loss or, more effectively, a Binary 
    Cross Entropy loss adapted from vision-language models--is sufficient for this 
    task. This simplifies the hyperparameter search and reduces computational 
    complexity compared to multi-loss frameworks.

    \item \textbf{High memory efficiency:} Through strategic batch construction 
    (sampling only 2 frames), optimized augmentation strategies, and the 
    aforementioned design choices, we achieve state-of-the-art accuracy with a 
    memory footprint of less than 1GB—an order of magnitude less than standard 
    contrastive methods.
\end{enumerate}

The remainder of this paper is structured as follows: Section 2 details 
our method, including batch construction, similarity computation, masking, 
and loss functions. Section 3 describes experimental setups, datasets, and 
baselines. Section 4 presents results and comparisons with state-of-the-art 
methods. Section 5 discusses related work, and Section 6 concludes with 
future directions.

%% file: method.tex
Our method is applied to video datasets of animals of the same type
where the bounding boxes of each animal have already been extracted.

\subsection{Training Loop}

The schematic diagram of the training loop of our self-supervised method 
is displayed in Figure~\ref{fig:training_loop}. 

\subsubsection{Construct a training batch}

To construct a training batch, we randomly select a user-defined number 
of frames, $K$ (where $K \ge 2$), denoted as $\{F_1, F_2, \dots, F_K\}$.
While generalised for K frames, setting $K = 2$ minimises the memory needed
for training. Our diagram is a special case where $K = 3$. We then crop 
the corresponding bounding boxes to each frame. Each cropped image then 
gets augmented twice, which doubles the number of images in the batch. 

\subsubsection{Compute the Similarity Matrix}

After the batch is constructed, it gets passed through a neural 
network, which converts each image to a high-dimensional feature 
vector. The batch of feature vectors is shown in 
Figure~\ref{fig:training_loop} as coloured blocks. The colours that 
contain the same combination of primary colours (e.g., purple and 
pink) indicate those two blocks represent two instances of augmentations 
applied to cropped images from the same frame. We then compute the 
similarity score for each and every other vector within the batch. 
The diagonal elements of the similarity matrix record 
the self-similarity for each vector.

\subsubsection{Construct the Mask for the Similarity Matrix}

The mask matrix has exactly the same size as the similarity matrix.
Colour red, green and white denote the similarity scores that 
should be discarded, maximised and minimised. 


The matrix is organized into four quadrants representing the similarity 
between the two augmented views (Instance 1 and Instance 2). As shown in 
Figure~\ref{fig:training_loop}, the diagonal quadrants (Top-Left, Bottom-Right) 
represent intra-view similarities, while off-diagonal quadrants represent 
inter-view similarities.

We then divide each quadrant into submatrices such that each submatrix 
contains the similarity score of feature vectors originated from each
frame. Each off-diagonal submatrix contains the similarity score between
feature vectors originated from different frames. Each diagonal submatrix
contains the similarity score between feature vectors originated from the same
frames. We discard the diagonal elements of the top-left and bottom-right
quadrants since they record the self-similarity score of each feature vector
originated from the same augmented instance. However, we mark the diagonal
elements of the bottom-left and top-right quadrants as green, since they
record the self-similarity score of each feature vector originated from
different augmented instances. To determine the mask for off-diagonal 
submatrices, we employ a heuristic based on the assumption that the 
network---initialized with a pre-trained backbone---is capable of 
extracting meaningful features from the outset. Consequently, we treat the 
optimal assignment of pairs (the permutation that maximizes total similarity) 
as a proxy for ground truth. We apply the Hungarian Algorithm to each 
off-diagonal submatrix to identify these positive pairs. This process 
allows the network to self-bootstrap: the pre-trained features guide 
the initial masking, which in turn supervises the network to learn 
more discriminative representations specific to the animal identities.

\subsubsection{Loss Function}

\begin{figure}[h]
    \centering
    \includegraphics[width=\textwidth]{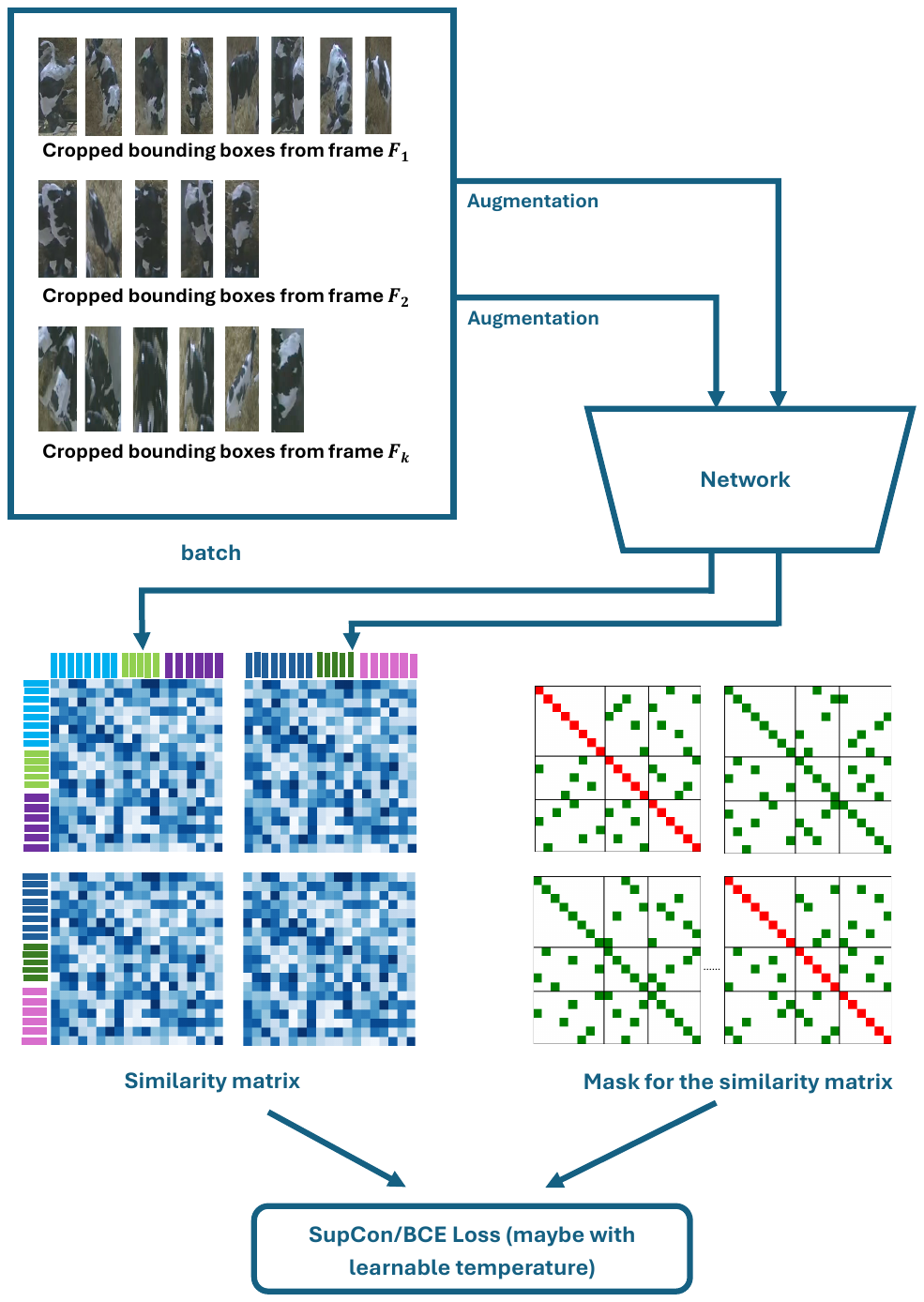}
    \caption{Schematic Diagram of the training pipeline of our method.}
    \label{fig:training_loop}
\end{figure}

We feed both the similarity matrix and the mask to the loss function. The
choice of the loss function could either be a Supervised Contrastive Loss, 
where the total loss, $\mathcal{L}_{\text{SupCon}}$, is the average of the 
loss of each positive in the row of each similarity matrix (as 
Equation~\ref{eq:total_supcon_loss} shows):
\begin{equation}
    \mathcal{L}_{\text{SupCon}}=\frac{1}{|\mathcal{P}|}\sum_{(i,j\in \mathcal{P})}
    \mathcal{L}_{i,j}
    \label{eq:total_supcon_loss}
\end{equation}
Where $\mathcal{P}$ denotes the set of matrix index pairs that 
are marked by colour green in the corresponding mask (as 
Figure~\ref{fig:training_loop} shows), and $i$ and $j$ denote
the row and column numbers respectively. The loss of each positive is
described by Equation~\ref{eq:supcon_loss_per_positive}:

\begin{equation}
    \mathcal{L}_{i,j} = -\log \frac{\exp(\mathbf{Sim}_{i,j}/ \tau)}
    {\sum_{k\in\{1:N\}, k\neq i} \exp(\mathbf{Sim}_{i,k}/ \tau)}
    \label{eq:supcon_loss_per_positive}
\end{equation}
Where $\mathbf{Sim}$ is the similarity matrix, $N$ is the size of
the similarity matrix, $\tau$ is a temperature parameter that can 
either be arbitrarily set by the user to a constant or a learnable 
parameter and $k$ is the column number. $k\neq i$ indicates that 
the diagonals marked as red ($i\neq j$) in Figure~\ref{fig:training_loop} 
are discarded.

The loss could also be the Binary Cross Entropy Loss, where the total 
loss, $\mathcal{L}_{\text{BCE}}$, is computed as 
\begin{equation}
\mathcal{L}_{\text{BCE}}
=
- \frac{1}{N^2}
\sum_{i,j\in\{1:N,1:N\},  i\neq j}
\log \sigma\!\left(
M_{ij} \, \frac{\mathbf{Sim}_{i,j}}{\tau}
\right)
\label{eq:bce_loss}
\end{equation}
Where $\sigma$ is the sigmoid function and $M$ is the mask matrix 
in Figure~\ref{fig:training_loop} where colour green indicates 
$M_{ij}=1$ and colour white indicates $M_{ij}=-1$. Like the SupCon 
loss, the diagonals marked as red are discarded. While $\tau$ can 
be treated as a fixed hyperparameter, we treat it as a learnable 
parameter to improve training stability. Following the strategy in 
SigLIP\cite{siglip}, we parameterize the logit scaling as 
$\mathbf{Sim}_{i,j} \cdot t + b$, where $t$ and $b$ are learnable scalars 
initialized to 10 and -10. We clamped the value of $t$ between $0-100$.

\subsubsection{Updating the Network}

Once the loss is computed, we use backpropagation to compute the gradient,
then use SGD with momentum to update the network.

\subsection{Inference}

Once the network is trained, we convert all images from the dataset to 
feature vectors by forward passing them through the network. We then use 
the KMeans algorithm to group the vectors into $N_{ID}$ clusters, where 
$N_{ID}$ corresponds to the known number of distinct animal identities 
in the video. Throughout the entire training and inference process, no 
ground truth identity of each cropped image is used.

%% file: experimental_details.tex

\subsection{Choice of Datasets}

We applied our method to datasets constructed from 3 separate videos. One is the 
8-Calves dataset, constructed from a 1-hr video of 8 calves feeding in a barn. 
The other two are selected from the 59 videos in the 3D-POP\cite{3DPOP} dataset that records 
the feeding of 18 pigeons on different days. To maximise the robustness of our 
method, we selected the two videos containing the highest number of pigeons (10) 
while sharing the fewest common identities (4). These videos were recorded by 
different cameras on different days, ensuring the visual data is as distinct 
as possible. This selection strategy evaluates whether our method generalizes 
to videos taken from different camera angles on different days that record a 
different set of individuals. The specific sequences used were 
Sequence11\_n10\_01072022-Cam2 and Sequence19\_n10\_13072022-Cam4 (hereafter referred 
to as Seq11 and Seq19). We used the pre-existing bounding boxes provided by these 
datasets to crop the frames and constructed 3 image datasets.

\subsection{Network Design}

We want to use a network where the GPU memory consumption during training is
minimal. We tested 2 network setups. The first setup contains a pre-trained 
backbone with its classification module replaced by a linear layer with an
output size of 64. Every parameter in this network is trainable. The second 
setup contains a frozen pre-trained network with an MLP that has the following 
structure:
\begin{itemize}
    \item Linear(backbone\_output\_size, 256) + BatchNorm1d + ReLU
    \item Linear(256, 128) + BatchNorm1d + ReLU
    \item Linear(128, 128) + BatchNorm1d + ReLU
    \item Linear(128, 64)
\end{itemize}
For the first setup, we limit the choice of backbone to ResNet18 and EfficientNet-B0; 
we specifically avoided training transformer backbones to avoid the high memory 
overhead associated with attention mechanisms. For the 2nd setup, we expanded
the choice of backbone to ConvNextV2, since the backbone would be frozen. 
Implementation-wise, all methods were developed using PyTorch. We utilized 
pre-trained weights from the \texttt{torchvision} library for ResNet18 and 
EfficientNet-B0, while ConvNextV2 weights were sourced from Hugging Face.

\subsection{Training Details}

\textbf{Augmentation:} We applied two augmentations to double the size 
of our batch. They are random horizontal flip and random resized crop to images 
with a size of (224, 224) and a scale of (0.2, 1.0). Unlike standard contrastive
learning approaches such as simCLR\cite{simclr} and MoCo\cite{MoCo}, we did not 
apply color jitter or Gaussian blur.

\textbf{Sampling Strategy (Batch Size):} To minimise GPU memory consumption, 
we constructed each training batch by randomly sampling $K=2$ frames from 
the video. Since the number of visible bounding boxes varies due to occlusion or 
detection failures, the effective batch size is dynamic. For a video with a maximum 
of 10 individuals, the effective batch size is $\le 40$ image crops per iteration 
($K=2 \times 10 \text{ max animals} \times 2 \text{ augmented views}$).

\textbf{Loss Function:} We tested 3 loss functions: BCE with a learnable 
temperature (as Equation~\ref{eq:bce_loss} shows), SupCon with a fixed temperature, 0.5, and 
SupCon with a learnable temperature. For the SupCon with a learnable temperature, we replaced $1/\tau$ in
Equation~\ref{eq:supcon_loss_per_positive} with a weight, $t$. Followed the strategy
of CLIP\cite{CLIP}, we clamped the value of $t$ between $0-100$ and set its initial 
value to 14.

\textbf{Optimiser Settings:} We used SGD to update the network weights of our method.
We set momentum to 0.9, weight decay to 0 and an initial learning rate to 
$0.3 \times batch size / 256$. We also used a cosine annealing scheduler to 
alter the learning rate during training.

\subsection{Inference Details}

We used KMeans to group the vectors into clusters. Then to calculate the 
accuracy of our method, we optimally matched the clusters with the ground 
truth identities via the Hungarian Algorithm.

\subsection{Baselines for Comparison}

To evaluate the effectiveness of our approach against standard contrastive learning 
baselines, we compared our method against supervised and self-supervised methods. 
We implemented all methods in PyTorch using a ResNet18 backbone.

For self-supervised methods, we tested SimCLR\cite{simclr} and MoCo\cite{MoCo}. 
For MoCo, the momentum encoder was also a ResNet18 instance with gradient updates 
disabled. For MoCo, we utilized a momentum value of $m=0.999$ and a memory bank 
size 32 times of the batch size. Crucially, to test the feasibility of these 
methods on the same resource-constrained hardware targeted by our approach, we 
limited their batch size to 256. While these methods typically utilize larger 
batches, this constraint allows for a direct comparison of resource efficiency.
We kept the Optimiser settings for these two methods identical to our own.

For the supervised approach, the network architecture, hyperparameters and
batch construction were all kept the same as our own method. The key distinctions
are that, (1) the dataset was split into train, validation and test sets; (2) 
During training, no similarity matrix is computed. The ground truths were 
directly used to compute the Cross Entropy Loss. We also employed an early 
stopping condition based on the validation loss with patience of 10 epochs 
to prevent overfitting.

%% file: results.tex
\subsection{Comparison with simCLR and MoCo}

While standard contrastive methods like SimCLR and MoCo are powerful, 
our results demonstrate they are inefficient for this specific domain. 
As shown in Table 1, these methods require over 10GB of memory per batch 
yet fail to achieve meaningful accuracy ($\sim$ 30\%).

Conversely, our method achieves state-of-the-art accuracy ($>$97\%) with a 
memory footprint of less than 1GB. This efficiency offers a dual advantage:

\begin{enumerate}
    \item Accessibility: It enables deployment on lower-end hardware.
    \item Scalability: Even on a high-end GPU, our method frees up GPU
memory, providing users the options to run other computational tasks on
the same GPU concurrently.
\end{enumerate}

\subsection{Comparison with the SOTA on 3D-POP and 8-Calves}

While 3D-MuPPET achieves state-of-the-art performance on the 3D-POP
dataset (0.96 ID-F1), it relies on a sequential tracking paradigm 
(using SORT on extracted keypoints). A fundamental limitation of 
sequential trackers is error propagation: a single identity switch 
can persist indefinitely, requiring manual intervention to correct.
This limitation becomes acute in long-duration footage. For instance, 
when applying standard tracking baselines (ByteTrack/Bot-SORT on 
bounding boxes) to the 1-hour 8-Calves dataset, we observed a 
performance drop to an ID-F1 of approximately 0.15.

In contrast, our method treats identification as a global clustering 
problem rather than a sequential tracking problem. By grouping feature 
vectors from the entire video simultaneously (as described in the 
Method Section), our approach eliminates the dependency on previous 
frames, thereby preventing error propagation. This allows our method 
to maintain high accuracy ($>$ 90\%) on long-duration videos where 
sequential trackers fail.

\subsection{Comparison with Supervised Baselines}

Remarkably, our self-supervised method (using BCE) matches or exceeds the 
performance of the supervised baseline trained on 1,000 labeled frames.
For example, on the 3D-POP Sequence19 dataset, the supervised baseline achieves 
97\% accuracy only after training on 1,000 frames. In contrast, our method 
achieves 99.8\% accuracy  without utilizing any ground-truth identity labels 
during training. This highlights a significant practical advantage: to replicate 
the performance of our self-supervised pipeline using a standard supervised 
approach, a user would be required to manually annotate at least 1,200 frames 
(1,000 for training and 200 for validation). Our method eliminates this manual 
bottleneck entirely while maintaining state-of-the-art accuracy.

\subsection{Ablation Studies on our method}

\subsubsection{Comparison of Loss functions}

We evaluated the impact of the loss function on model performance. As shown 
in Tables~\cref{tab:results_8calves,tab:results_seq11,tab:results_seq19}, the 
Binary Cross Entropy (BCE) loss consistently outperforms or matches the 
Supervised Contrastive (SupCon) loss. For instance, on Sequence 11, BCE 
achieves an accuracy of 97.6\% compared to 92.6\% for SupCon with a fixed 
temperature. While introducing a learnable temperature to SupCon closes this 
gap, BCE remains the most robust and stable choice across all datasets without 
requiring additional hyperparameter tuning.

\subsubsection{Memory Usage vs Accuracy}

To further investigate the lower bound of GPU memory consumption during training,
we experimented with freezing the backbone and training only the projection MLP. 
This strategy yields significant memory savings. For ResNet18, freezing 
the backbone reduces memory usage by approximately 60\% (from 1.04 GB to 
0.42 GB). Despite this massive reduction in computational cost, the model retains 
strong performance, achieving between 86--90\% accuracy across datasets. This 
confirms that a pre-trained ImageNet backbone extracts sufficiently discriminative 
features for animal identification, allowing for extremely lightweight adaptation.

\subsubsection{Backbone Architecture Analysis}

We also analyzed the choice of backbone architecture. Contrary to expectations, 
EfficientNet-B0 proved less memory-efficient than ResNet18 for our specific 
training setup. Despite having fewer parameters, EfficientNet-B0 requires 
storing significantly more activations during the forward pass due to its depth, 
resulting in a memory footprint of 3.44 GB per batch compared to 1.04 GB for 
ResNet18. Consequently, ResNet18 remains the optimal choice for minimizing peak 
memory usage.

\begin{table}[h]
\footnotesize
\centering
\begin{tabular}{ccccc}
\toprule
\textbf{Loss} & \textbf{Network} & \makecell{\textbf{Train} \\ \textbf{MLP Only}} & \makecell{\textbf{Memory Per} \\ \textbf{Batch (GB)}} & \textbf{Accuracy}  \\
\midrule
BCE & ResNet18 & False & 1.044 $\pm$ 0.000 & 0.976 $\pm$ 0.021 \\
BCE & ResNet18 & True & 0.418 $\pm$ 0.000 & 0.866 $\pm$ 0.008 \\
BCE & ConvNextV2 & True & 0.696 $\pm$ 0.000 & 0.899 $\pm$ 0.003 \\
BCE & EfficientNet-B0 & False & 3.440 $\pm$ 0.001 & 0.980 $\pm$ 0.011 \\
BCE & EfficientNet-B0 & True & 0.521 $\pm$ 0.000 & 0.897 $\pm$ 0.079 \\
Supcon & ResNet18 & False & 1.044 $\pm$ 0.000 & 0.926 $\pm$ 0.036 \\
Supcon & ResNet18 & True & 0.418 $\pm$ 0.000 & 0.643 $\pm$ 0.047 \\
Supcon & ConvNextV2 & True & 0.696 $\pm$ 0.000 & 0.740 $\pm$ 0.070 \\
Supcon\textsuperscript{*} & ResNet18 & False & 1.044 $\pm$ 0.000 & 0.977 $\pm$ 0.012 \\
Supcon\textsuperscript{*} & ResNet18 & True & 0.418 $\pm$ 0.000 & 0.736 $\pm$ 0.008 \\
Supcon\textsuperscript{*} & ConvNextV2 & True & 0.696 $\pm$ 0.000 & 0.715 $\pm$ 0.066 \\
MoCo & ResNet18 & False & 7.369 $\pm$ 0.003 & 0.278 $\pm$ 0.028 \\
SimCLR & ResNet18 & False & 10.867 $\pm$ 0.004 & 0.312 $\pm$ 0.050 \\
\bottomrule
\end{tabular}
\caption{3D-POP Sequence11 Epoch 10. The asterisk (*) indicates 
models trained using Supcon with learnable temperature.}
\label{tab:results_seq11}
\end{table}

\begin{table}[h]
\footnotesize
\centering
\begin{tabular}{ccccc}
\toprule
\textbf{Loss} & \textbf{Network} & \makecell{\textbf{Train} \\ \textbf{MLP Only}} & \makecell{\textbf{Memory Per} \\ \textbf{Batch (GB)}} & \textbf{Accuracy}  \\
\midrule
BCE & ResNet18 & False & 1.044 $\pm$ 0.000 & 0.998 $\pm$ 0.000 \\
BCE & ResNet18 & True & 0.418 $\pm$ 0.000 & 0.898 $\pm$ 0.089 \\
BCE & ConvNextV2 & True & 0.696 $\pm$ 0.000 & 0.847 $\pm$ 0.052 \\
BCE & EfficientNet-B0 & False & 3.441 $\pm$ 0.000 & 0.993 $\pm$ 0.003 \\
BCE & EfficientNet-B0 & True & 0.521 $\pm$ 0.000 & 0.939 $\pm$ 0.084 \\
Supcon & ResNet18 & False & 1.044 $\pm$ 0.001 & 0.787 $\pm$ 0.088 \\
Supcon & ResNet18 & True & 0.418 $\pm$ 0.000 & 0.716 $\pm$ 0.017 \\
Supcon & ConvNextV2 & True & 0.696 $\pm$ 0.000 & 0.747 $\pm$ 0.055 \\
Supcon\textsuperscript{*} & ResNet18 & False & 1.043 $\pm$ 0.002 & 0.995 $\pm$ 0.005 \\
Supcon\textsuperscript{*} & ResNet18 & True & 0.418 $\pm$ 0.000 & 0.854 $\pm$ 0.120 \\
Supcon\textsuperscript{*} & ConvNextV2 & True & 0.696 $\pm$ 0.000 & 0.823 $\pm$ 0.022 \\
MoCo & ResNet18 & False & 7.367 $\pm$ 0.006 & 0.245 $\pm$ 0.087 \\
SimCLR & ResNet18 & False & 10.872 $\pm$ 0.001 & 0.299 $\pm$ 0.023 \\
\bottomrule
\end{tabular}
\caption{3D-POP Sequence19 Epoch 10. The asterisk (*) 
indicates models trained using Supcon with learnable temperature.}
\label{tab:results_seq19}
\end{table}

\begin{table}[h]
\footnotesize
\centering
\begin{tabular}{ccccc}
\toprule
\textbf{Loss} & \textbf{Network} & \makecell{\textbf{Train} \\ \textbf{MLP Only}} & \makecell{\textbf{Memory Per} \\ \textbf{Batch (GB)}} & \textbf{Accuracy}  \\
\midrule
BCE & ResNet18 & False & 0.871 $\pm$ 0.001 & 0.981 $\pm$ 0.013 \\
BCE & ResNet18 & True & 0.355 $\pm$ 0.000 & 0.891 $\pm$ 0.022 \\
BCE & ConvNextV2 & True & 0.580 $\pm$ 0.000 & 0.902 $\pm$ 0.014 \\
BCE & EfficientNet-B0 & False & 2.791 $\pm$ 0.003 & 0.992 $\pm$ 0.004 \\
BCE & EfficientNet-B0 & True & 0.435 $\pm$ 0.000 & 0.932 $\pm$ 0.001 \\
Supcon & ResNet18 & False & 0.871 $\pm$ 0.000 & 0.978 $\pm$ 0.014 \\
Supcon & ResNet18 & True & 0.355 $\pm$ 0.000 & 0.809 $\pm$ 0.021 \\
Supcon & ConvNextV2 & True & 0.580 $\pm$ 0.001 & 0.823 $\pm$ 0.062 \\
Supcon\textsuperscript{*} & ResNet18 & False & 0.872 $\pm$ 0.000 & 0.986 $\pm$ 0.014 \\
Supcon\textsuperscript{*} & ResNet18 & True & 0.355 $\pm$ 0.000 & 0.900 $\pm$ 0.022 \\
Supcon\textsuperscript{*} & ConvNextV2 & True & 0.580 $\pm$ 0.000 & 0.907 $\pm$ 0.028 \\
MoCo & ResNet18 & False & 7.371 $\pm$ 0.001 & 0.220 $\pm$ 0.049 \\
SimCLR & ResNet18 & False & 10.870 $\pm$ 0.003 & 0.213 $\pm$ 0.009 \\
\bottomrule
\end{tabular}
\caption{8-calves Epoch 2. The asterisk (*) indicates models trained using Supcon 
with learnable temperature.}
\label{tab:results_8calves}
\end{table}

\begin{table}[h]
\footnotesize
\centering
\begin{tabular}{ccccccc}
\toprule
\textbf{Dataset} & \textbf{Network} & \makecell{\textbf{Train} \\ \textbf{MLP Only}} & \makecell{\textbf{Train} \\ \textbf{Frames}} & \makecell{\textbf{Val} \\ \textbf{Frames}} & \makecell{\textbf{Memory Per} \\ \textbf{Batch (GB)}} & \textbf{Accuracy} \\
\midrule
3D-POP Sequence11 & EfficientNet-B0 & True & 1000 & 200 & 0.31 & 0.77 \\
3D-POP Sequence11 & ResNet18 & False & 100 & 100 & 0.66 & 0.88 \\
3D-POP Sequence11 & ResNet18 & False & 1000 & 200 & 0.66 & 0.96 \\
3D-POP Sequence19 & EfficientNet-B0 & True & 1000 & 200 & 0.31 & 0.72 \\
3D-POP Sequence19 & ResNet18 & False & 100 & 100 & 0.66 & 0.89 \\
3D-POP Sequence19 & ResNet18 & False & 1000 & 200 & 0.66 & 0.97 \\
8-calves & EfficientNet-B0 & True & 1000 & 200 & 0.27 & 0.61 \\
8-calves & ResNet18 & False & 100 & 100 & 0.57 & 0.89 \\
8-calves & ResNet18 & False & 1000 & 200 & 0.57 & 0.96 \\
\bottomrule
\end{tabular}
\caption{Supervised baseline results on 3D-POP and 8-calves datasets.}
\label{tab:supervised_baselines}
\end{table}

%% file: related_work.tex


The most similar work to ours is the ContrasT\cite{Plaen_2024} paper. It 
also uses random frame sampling and the application of the Hungarian 
Algorithm to assign pseudo-labels during training. However, key distinctions 
arise from our focus on resource efficiency and single-video identification. 
While ContrasT constructs batches by sampling frames from multiple videos 
to maximize negative diversity, our method samples frames solely from a single 
video to minimize computational overhead. Whilst ContrasT requires fine-tuning 
the entire transformer backbone, ours functions effectively even when fine-tuned
with a frozen backbone. And finally, whilst ContrasT employs multiple loss functions 
to handle joint detection and tracking, ours only requires only a single loss 
function (SupCon or BCE), and operates on lower-resolution inputs to further
lower the GPU memory footprint.

Our loss function design draws inspiration from vision-language models 
like CLIP and SigLIP, specifically in the use of learnable temperature 
parameters and the adaptation of Binary Cross Entropy (BCE) for contrastive 
tasks. The main differences between our method and CLIP and SigLIP are the 
input modalities and training objectives. CLIP and SigLIP rely on pairing 
images with text and training from scratch on massive datasets. Our method, 
conversely, operates purely in the visual domain using image--image pairs
for self-supervision and leverages pre-trained weights to bootstrap performance. 
Additionally, we employ the Hungarian Algorithm to dynamically resolve 
positive and negative pairs within the batch, a step absent in the static 
image-text pairing of CLIP and SigLIP.

While standard self-supervised frameworks like SimCLR and MoCo 
inform our contrastive approach, they are ill-suited for the specific 
constraints of our domain. These methods typically require training from 
scratch and rely on large memory banks (as in MoCo) or massive batch sizes 
with random sampling to learn discriminative features. Our experiments 
demonstrate that these requirements lead to excessive memory consumption 
($>$10GB) with suboptimal accuracy for animal identification tasks. Our 
method mitigates this by employing a structured sampling strategy (pairing 
frames from the same video) rather than random batch sampling, and by 
utilizing augmentations to double the effective batch size, a feature shared 
with SimCLR but absent in MoCo's standard implementation. This allows us 
to achieve high accuracy with a memory footprint of less than 1GB.

%% file: conclusion.tex
We have presented a highly efficient, self-supervised method for identifying 
individual animals within long-duration video sequences, operating under the 
practical constraint of a known, fixed population size. By reframing the problem 
as a global clustering task rather than a sequential tracking problem, our 
approach eliminates the fundamental issue of error propagation that plagues 
frame-by-frame association methods, enabling robust performance over extended 
periods.

The core of our method lies in a memory-conscious design that samples pairs of 
frames, employs a self-bootstrapping mechanism using the Hungarian algorithm 
for in-batch pseudo-label assignment, and utilizes a simple yet effective loss 
function—adapting the Binary Cross Entropy from vision-language models. This 
design allows us to leverage pre-trained visual backbones effectively, often 
in a frozen state, reducing the trainable parameters to a lightweight projection 
head.

Our experiments on challenging real-world datasets (3D-POP pigeons and 8-calves) 
demonstrate that this minimalist framework achieves state-of-the-art accuracy 
($>$97\%), matching or surpassing supervised baselines that require over 1,000 
manually annotated frames, while consuming less than 1GB of GPU memory per 
training batch. This represents an order-of-magnitude reduction in memory 
compared to standard contrastive learning approaches (SimCLR, MoCo), making 
high-quality animal identification accessible on consumer-grade hardware.

The key practical outcome is the removal of the manual annotation bottleneck 
for a common class of wildlife and livestock monitoring scenarios. Researchers 
can now obtain accurate individual identities from long videos without any 
labeled data, relying only on pre-existing detections and the prior knowledge 
of group size.

Future work will focus on extending this framework to more dynamic settings, 
such as open-world scenarios where the number of individuals is unknown, and 
on integrating temporal consistency models to further refine features across 
very long sequences. The principles of resource-efficient, task-specific 
self-supervision established here could also benefit other domains where 
labeled data is scarce and computational resources are limited.